\documentclass[final,5p,times,twocolumn]{elsarticle}

\usepackage{amssymb}
\usepackage{amsmath}
\usepackage{xcolor}
\usepackage{graphicx}
\usepackage{amsfonts}
\usepackage{makecell}
\usepackage{subcaption}
\usepackage{hyperref}

\usepackage{amssymb}
\usepackage{amsmath}

\usepackage[T1]{fontenc}
\usepackage[utf8]{inputenc}

\journal{IFAC Journal of Systems and Control}

\begin{document}

\begin{frontmatter}



\title{Generalizing deep reinforcement learning across cable-driven parallel robot configurations with actuator-level policies}


\author[label1,label2]{Abir Bouaouda}
\author[label1,label2,label3]{Mohamed Boutayeb}
\author[label2]{Fran\c{c}ois Charpillet}
\author[label4]{Dominique Martinez}
\author[label1]{R\'emi Pannequin}
\affiliation[label1]{organization={Universit\'e de Lorraine, CRAN},
            city={Nancy},
            postcode={F-54000},
            country={France}}
\affiliation[label2]{organization={Universit\'e de Lorraine, INRIA, CNRS, LORIA},
            city={Nancy},
            postcode={F-54000},
            country={France}}
\affiliation[label3]{organization={International University of Rabat},
            city={Rabat},
            postcode={11103},
            country={Morocco}}
\affiliation[label4]{organization={Aix-Marseille Universit\'e, CNRS, ISM},
            city={Marseille},
            postcode={F-13009},
            country={France}}
\ead{firstname.lastname@univ-lorraine.fr}
\ead{firstname.lastname@inria.fr}
\ead{firstname.lastname@univ-amu.fr}

\begin{abstract}
Cable-driven parallel robots (CDPRs) present diverse configurations and complex control challenges, which can be addressed by deep reinforcement learning (DRL) by learning their nonlinear dynamics. However, DRL methods often require extensive training time, and the resulting policies do not generalize well to different robot configurations or varying numbers of actuators. In this article, we introduce a novel DRL approach for controlling CDPRs that does not depend on the specific robot configuration. Our method trains an actuator-level policy that controls each motor to achieve its target cable length, in contrast to conventional DRL approaches that learn to control the entire robot to reach a desired end-effector position. To the best of our knowledge, this is the first work to apply DRL to control CDPRs using an actuator-level policy. This approach offers two main advantages: (i) a single shared policy can be applied to any CDPR configuration, regardless of actuator count, and (ii) reliance on inverse kinematics, avoiding the more challenging forward kinematics problem. Training is performed in simulation, and the learned policy is successfully transferred to a real CDPR. Experimental results show that the actuator-level policy (ALP) surpasses traditional reinforcement learning methods in both robustness and precision. We further control a real 8-motor CDPR with 3D motion using a policy trained on a simulated 4-motor planar CDPR operating in 2D. This illustrates that the proposed method is applicable to any CDPR configuration, independent of actuator number or placement.
\end{abstract}

%


\end{frontmatter}



\section{Introduction}

Cable-driven parallel robots (CDPRs) have garnered considerable interest in recent years due to their wide range of applications, including medical rehabilitation \cite{surdilovic_string-man_2004}, manufacturing assembly tasks \cite{pott_large-scale_nodate}, construction \cite{williams_contour-crafting-cartesian-cable_2008}, and logistics for sorting and packaging \cite{lamaury_control_2013}. This diversity of applications brings a variety of control challenges, for which Deep Reinforcement Learning (DRL) has emerged as a promising solution, thanks to its ability to capture the complex, non-linear dynamics involved in CDPR control. Some researchers have combined DRL with classical control strategies, while others have relied solely on DRL. For example, in \cite{ma_control_2019,sancak_position_2022}, the authors employ the DDPG algorithm to learn a control policy for a CDPR, where the resulting policy is represented by a neural network that takes the robot's state as input and outputs the torques to be applied to its actuators.

However, the use of DRL for controlling CDPRs is not without challenges. One major difficulty is the significant training time required to obtain an effective policy. Moreover, with current state-of-the-art methods, the learned policy often cannot be applied to robots with a different number of actuators, which is a major limitation since the number of actuators may vary between applications. Even when the actuator count remains the same, the policy may not generalize well to different robot configurations, such as changes in mass or actuator positions.

Several approaches have been proposed to address this limitation, including domain randomization \cite{tobin_domain_2017} and transfer learning \cite{rusu_sim--real_2018}. However, these methods are less effective when the input space itself changes.

A recent innovative approach to generalization, "Gato," was proposed by \cite{reed_generalist_2022}. Gato uses a single network with shared weights to perform a wide range of tasks, such as playing Atari, captioning images, chatting, and controlling a real robot arm, by deciding based on context whether to output text, joint torques, button presses, or other tokens. While promising, Gato requires a very large neural network and often needs fine-tuning for specific tasks, making it impractical for many applications. Furthermore, Gato achieves expert-level performance on only 180 out of 604 simulated control tasks.

While solving the generalization problem for large-scale control tasks, as attempted by Gato, may be an ultimate goal for the robotics community, it is also important to develop more efficient and practical solutions for specific classes of robots, such as Cable-Driven Parallel Robots (CDPRs). CDPRs are particularly suitable for studying generalization because their fixed-frame actuators allow for many possible configurations with varying numbers of actuators, resulting in different input spaces. This makes the generalization challenge both more difficult and more interesting to address.
\begin{figure}[thpb]
    \centering
    \framebox{\parbox{3in}{
        \centering
        \includegraphics[width=1\linewidth]{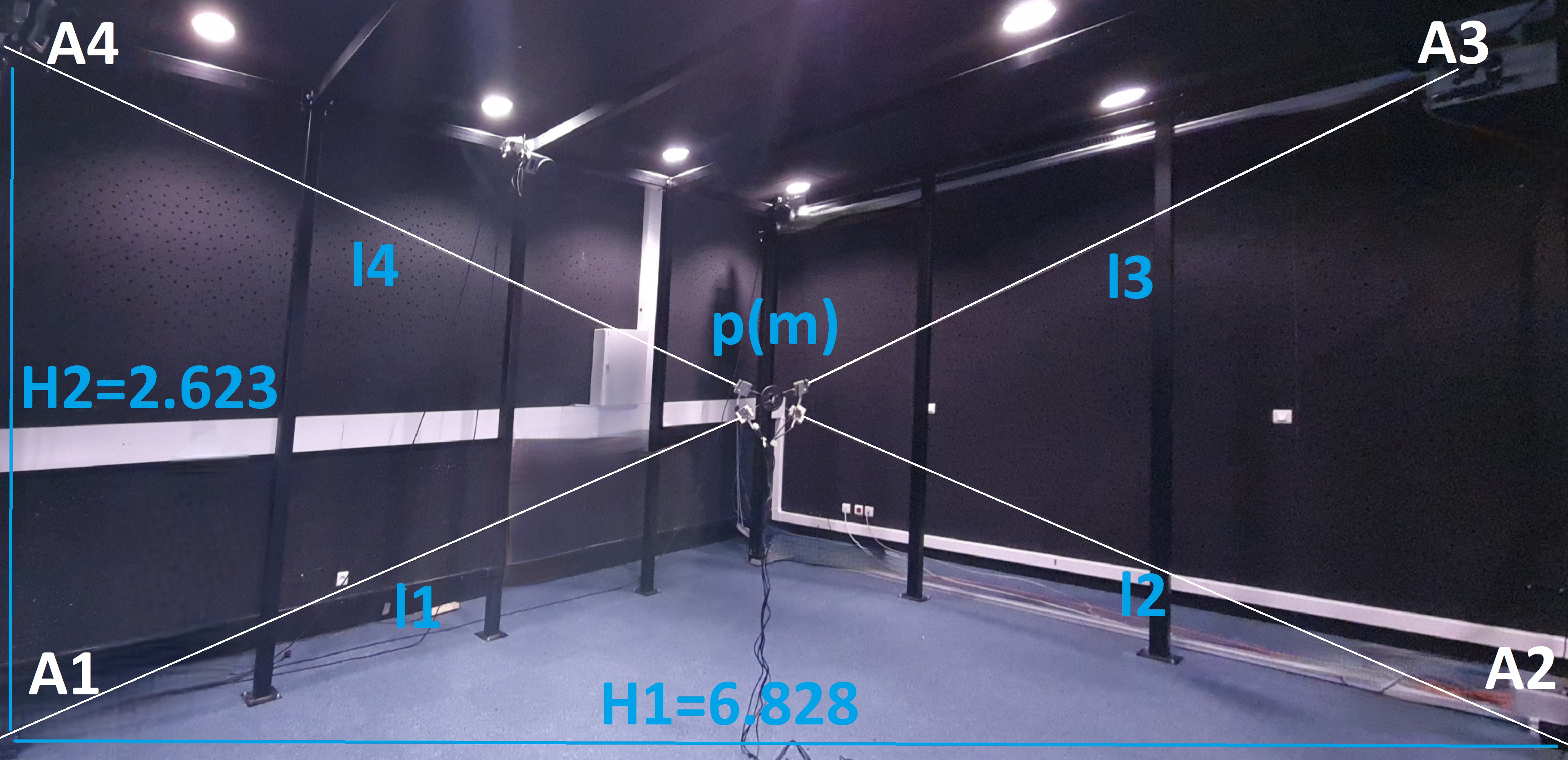}
    }}
    \caption{Two-DOF point-mass CDPR with four cables. Here, $l_k$ is the length of cable $k$, $A_k$ is the position of motor $k$, and $p$ is the position of the end-effector.}
    \label{fig:schemcablerobot}
\end{figure}
In this paper, we propose a new approach to address the generalization problem in the control of CDPRs. Our method is based on learning a policy at the actuator level rather than at the robot level. 
We name this approach "actuator-level policy" (ALP). It could be described as a multi-agent system in which each agent controls a single actuator. The learning is centralized, with a shared policy among all agents, but execution is decentralized, as each agent operates based on its own state and action. The only coordination occurs when computing the target length for each cable, which is determined using the robot's inverse kinematics and the desired end-effector position, ensuring all actuators share the same objective: moving the end-effector to the target.

The main advantage of this approach is that the learned policy can generalize to robots with different numbers and positions of actuators, since the policy is trained to control an actuator in any position, rather than relying on the full robot state. We evaluate our approach on simulated CDPRs with varying actuator counts and on a real robot with eight actuators. Results show that the ALP approach outperforms conventional reinforcement learning methods, as it enables control of a wide range of CDPRs using a single policy trained on a four-actuator robot (see Figure~\ref{fig:schemcablerobot}). Additionally, it surpasses classical control methods because it does not require direct measurement of the end-effector position—which is often unavailable without a motion capture system—and is more robust to errors in cable length measurements, since it does not depend on forward kinematics for position estimation.

\section{Materials and methods}
\begin{figure}[thpb]
    \centering
    \framebox{\parbox{3in}{
        \centering
        \resizebox{\linewidth}{!}{
            \fontsize{20pt}{11pt}\selectfont
            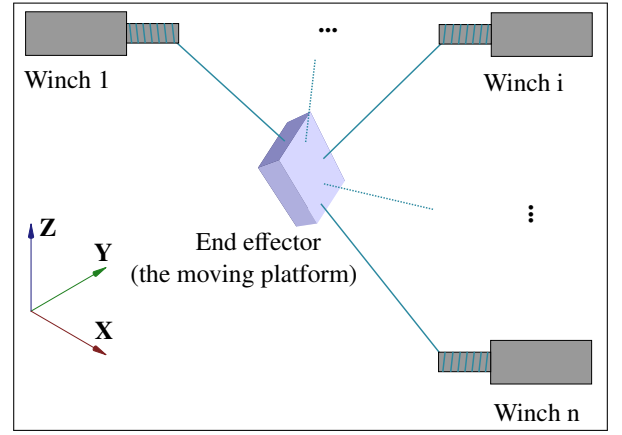
        }
    }}
\caption{CDPR with $n$ actuators. The end-effector (mobile platform) is connected to the base by $n$ cables, each actuated by a winch motor.}
    \label{fig:nCDPRs}
\end{figure}

\subsection{Cable driven parallel robots control problem formulation}
A CDPR is a type of parallel robot in which the end-effector is connected to the base by cables. These cables are actuated by motors, and the end-effector is moved by adjusting the lengths of the cables (see Figure~\ref{fig:nCDPRs}). The control problem for a CDPR consists of determining the torques to apply to the motors in order to move the end-effector to a desired 3D position.

Determining the cable lengths required for a given position and orientation of the mobile platform is known as the inverse kinematics (IK) problem. Assuming no cable sag, the cable lengths can be computed directly as the norms of the vectors connecting the motors to the end-effector. Conversely, determining the position and orientation of the mobile platform from the cable lengths is called the forward kinematics (FK) problem. While FK can be solved geometrically for translational CDPRs—using the intersection of spheres defined by the cables—it is more complex in the general case and is typically addressed using numerical methods.

In addition to solving the two problems, a model of the robot is needed to simulate its dynamics and to train the control policy. Developing such a model is beyond the scope of this paper; instead, we assume a black-box model that takes the desired motor speeds and the current state of the robot as input and outputs the next state. The model used in this research replicates the behavior of the real robot. For more details on robot modeling, we refer the reader to \cite{pott_cable_driven_2018}. The model outputs the robot state, including cable lengths, motor speeds, and motor currents. Using the cable lengths, the position and orientation of the mobile platform can be computed via the robot's forward kinematics.

Having defined the control problem, we now present how it can be addressed using reinforcement learning.
\section{Classical reinforcement learning controller for cdprs}
\begin{figure}[thpb]
    \centering
    \framebox{\parbox{3in}{
        \centering
        \resizebox{\linewidth}{!}{
            \fontsize{16pt}{11pt}\selectfont
            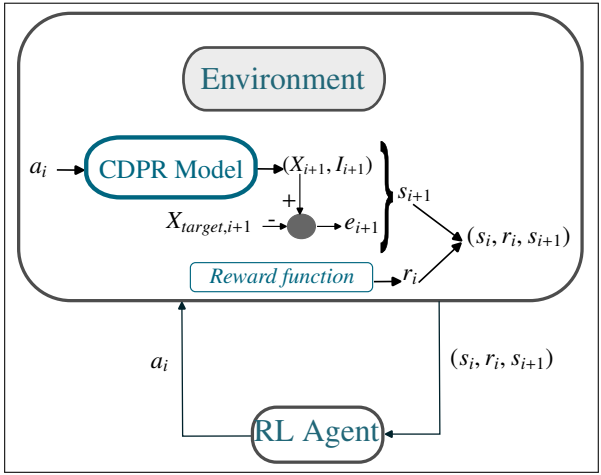
        }
    }}
    \caption{Interaction between the agent and the reinforcement learning environment. The agent receives the tuple $(s_t, a_t, r_t, s_{t+1})$ and outputs the action $a_t$ to be applied to the motors.}
    \label{fig:rlenv}
\end{figure}

Reinforcement learning is a field of machine learning that enables an agent to learn how to control an environment by taking actions based on observed states and receiving rewards. The agent learns a policy that maps states to actions in order to maximize the cumulative reward. In the context of CDPR control, the agent learns a policy that maps the robot's state to control signals applied to the motors to move the end-effector to a desired position.

There are several reinforcement learning algorithms suitable for continuous state and action spaces. In this work, we use three: DDPG \cite{lillicrap_continuous_2019}, PPO \cite{schulman_proximal_2017}, and SAC \cite{haarnoja_soft_2019}, to compare both on-policy and off-policy approaches. DDPG and SAC are off-policy algorithms, while PPO is an on-policy algorithm. The main distinction is that DDPG and SAC use a replay buffer to learn the policy, whereas PPO learns only from the most recent experiences.

In conventional reinforcement learning, the state describes the entire robot (see Figure~\ref{fig:rlenv}), and is defined as:
      \(  s_t=(X_t, X_{t-1}, e_t, e_{t-1}, I_t) \)
where $X_t$ is the position of the end-effector, $e_t$ is the error between the current and target positions, and $I_t$ is the motor current. All observations and actions are normalized to the range $[-1,1]$.

\subsection{ALP reinforcement learning controller}

\begin{figure}[htpb]
    \centering
    \framebox{\parbox{3in}{
        \centering
        \resizebox{0.8\linewidth}{!}{
            \fontsize{16pt}{11pt}\selectfont
            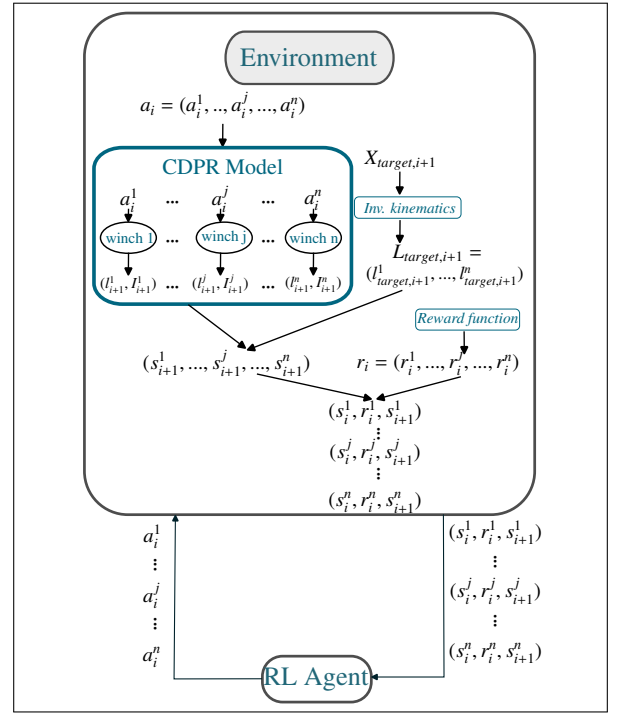
        }
    }}
    \caption{Interaction between the agent and the environment for ALP reinforcement learning. 
    At each step $t$, the agent interacts with the environment $n$ times, once for each actuator. 
From each interaction, the agent collects the tuple $(s_t^j, a_t^j, r_t^j, s_{t+1}^j)$ and outputs the action $a_t^j$ to be applied to motor $j$. The agent learns a single policy that maps the state of an actuator to the torque to be applied to its motor.}
    \label{fig:multirlenv}
\end{figure}

The main idea of ALP is to learn a policy that maps the state of each actuator to the torque applied to its motor, rather than mapping the full robot state to all motor torques. The key advantage of this approach is that the learned policy can generalize to robots with different numbers of actuators, since the policy is trained to control an individual actuator rather than the entire robot.

The agent-environment interaction for ALP reinforcement learning is illustrated in Figure~\ref{fig:multirlenv}.

The state of each actuator is defined as:
\[
s_t = (L_{t-3}^{j}, L_{t-2}^{j}, L_{t-1}^{j}, L_{t}^{j}, L_{t,\text{target}}^{j}, I_t^{j})
\]
where $L_{t}^{j}$ is the length of cable $j$ at step $t$, $L_{t,\text{target}}^{j}$ is the target length of cable $j$, and $I_t^{j}$ is the current of motor $j$. Including the cable length at step $t-3$ provides the agent with information about cable dynamics, tension at the other end, and cable speed, which are important for learning an effective policy (see \ref{sec:proofconcept} for more details).
All observations and actions are normalized to the range $[-1,1]$. Since there are $n$ actuators, at each step $t$, the agent performs $n$ interactions with the environment but learns a single shared policy. Thus, only one neural network is used to map the actuator state to the torque, and a single agent learns this policy from samples collected across all actuators.
\subsection{Reward-engineering}
\label{sec:reward}
The reward function is a crucial component of the reinforcement learning algorithm, guiding the agent to learn the optimal policy for a given task. For redundant CDPRs, multiple solutions can achieve the same target position. Therefore, a reward function based solely on position error is insufficient, as it may cause the policy to oscillate between solutions, resulting in noisy control. Furthermore, since our control signal is the motor speed (with a proportional control loop for motor current), the reward function must also account for the control signal to ensure the agent visits only states relevant for effective control.

\textbf{Error-based reward:} The first component of the reward function penalizes the position error:
       \( r_{1,t}=-\left\lVert e_t \right\rVert \)
where $e_t$ is the position error at step $t$, defined as the distance between the current and target positions of the end-effector. This term encourages the agent to minimize the position error.

\textbf{Energy optimization reward:} The second component penalizes the energy consumed by the motors, following the classical approach to energy optimization. Since energy consumption is related to motor current, we penalize the current as follows:
       \( r_{2,t}=-\left\lVert I_t \right\rVert^2 \)
where $I_t$ is the motor current at step $t$.

\textbf{Current limits reward:} The third component addresses current limits. Since we constrain the current using a saturation function (see \ref{sec:currentconstraints}), we encourage the agent to select actions that keep the current within safe bounds, as actions outside these limits will be saturated and thus ineffective. The reward is:
       \( r_{3,t}=- \left\lVert a_t-u_{sat,t}\right\rVert \)
where $a_t$ is the action chosen by the agent and $u_{sat,t}$ is the action after applying the saturation function.

So our final reward function is:
\begin{align}
    r_t  &= \alpha_1 r_{1,t} + \alpha_2 r_{2,t} + \alpha_3 r_{3,t} \\
    \begin{split}
        r_t =\ & -\alpha_1\left\lVert e_t \right\rVert 
        - \alpha_2\left\lVert I_t \right\rVert^2 
        - \alpha_3 \left\lVert a_t - u_{sat,t} \right\rVert
    \end{split} \label{eq:reward}
\end{align}
where $\alpha_1, \alpha_2, \alpha_3$ are coefficients that weight each term of the reward function. By tuning these coefficients, we adjust the importance of each component to achieve the best performance. After extensive testing, the best results were obtained with $\alpha_1=1$, $\alpha_2=0.001$, and $\alpha_3=0.5$. The energy optimization term ($\alpha_2$) improves efficiency, while the current limits term ($\alpha_3$) helps smooth the control signal, though it can slightly degrade tracking performance if set too high. Without this term, the control signal becomes too noisy for real robot use.

\subsection{Reward for ALP reinforcement learning}
The same reward function can be used for ALP reinforcement learning, where all actuators receive the same reward at each step. In this collaborative setting, an error on one actuator affects the reward for all, as the reward is based on the overall position error. We refer to this as reward 1 ($r_1$).

Alternatively, an individual reward can be used, based only on each actuator's own error:
\begin{align}
    \begin{split}
        r_t^j =\ & -\alpha_1\left| e_t^j \right| 
        - \alpha_2\left| I_t^j \right|^2 
        - \alpha_3 \left| a_t^j - u_{sat,t}^j \right|
    \end{split}
\end{align}
where $e_t^j$ is the error between the current and target lengths of cable $j$ at step $t$, $I_t^j$ is the current of motor $j$, $a_t^j$ is the action chosen by agent $j$, and $u_{sat,t}^j$ is the saturated action. The same coefficients $\alpha_1, \alpha_2, \alpha_3$ are used as in the collaborative reward. We refer to this as reward 2 ($r_2$).
In the training phase, we kept the same values for the coefficients $\alpha_1, \alpha_2, \alpha_3$.
\subsection{Target trajectories generation}

Visiting all states of the environment is crucial during training, as the agent must learn the optimal action for each possible state. In trajectory tracking tasks, the desired trajectory forms part of the state and can be generated as needed. However, it is important to ensure that generated trajectories are consistent with the robot's dynamic limits and sufficiently explore the entire state space.

\begin{figure}[thpb]
    \centering
    \framebox{\parbox{3in}{
        \centering
        \resizebox{\linewidth}{!}{
            \fontsize{11pt}{11pt}\selectfont
            \includegraphics{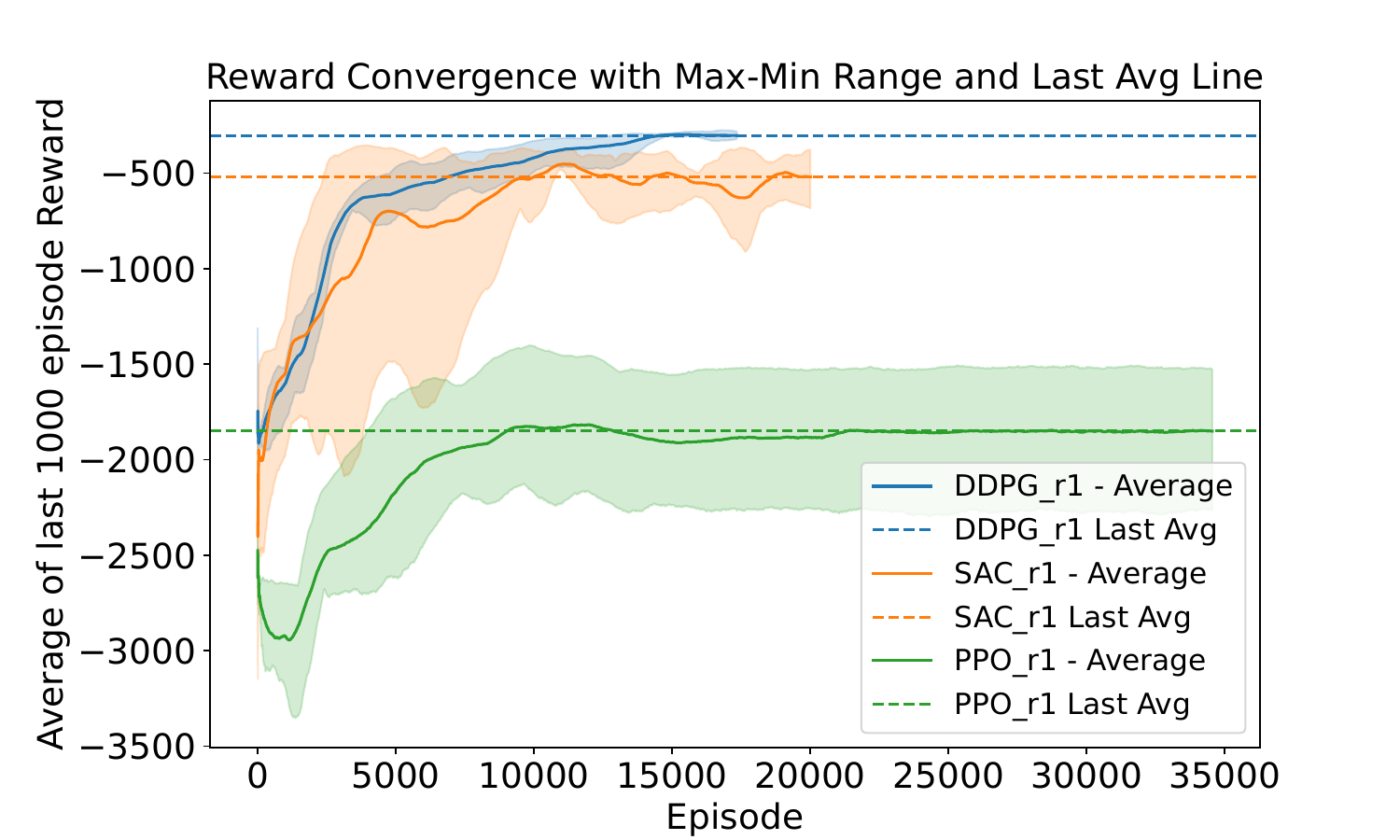}
        }
    }}
    \caption{Average reward of the ALP reinforcement learning approach. The solid curve shows the mean reward over four runs with different random seeds, while the shaded region represents the range (min–max) across runs. Each experiment was repeated four times with different random seeds.}
    \label{fig:reward}
\end{figure}
To address this, we generate random trajectories with random speeds and accelerations. Since the robot's workspace is bounded and there are constraints on speed and acceleration limits, it is essential to ensure that the generated trajectories comply with these constraints. For this purpose, we use the method described in \ref{sec:targetgen}.
\section{Experiments and results}
\subsection{Training setup}

The training was performed only on the 4-cable translational robot, assuming the mobile platform is a point mass and all cables are attached to the same point on the platform (see Figure~\ref{fig:schemcablerobot}). The robot parameters are provided in \ref{sec:hyperparams}, Table~\ref{tb:CDPRparams}. Training was conducted using three reinforcement learning algorithms: DDPG, PPO, and SAC, with hyperparameters detailed in \ref{sec:hyperparams}. Table~\ref{tb:Hyperparameters}.
\subsection{Training results for ALP}
\begin{figure}[thpb]
    \centering
    \framebox{\parbox{3in}{
        \centering
        \resizebox{\linewidth}{!}{
            \fontsize{11pt}{11pt}\selectfont
            \includegraphics{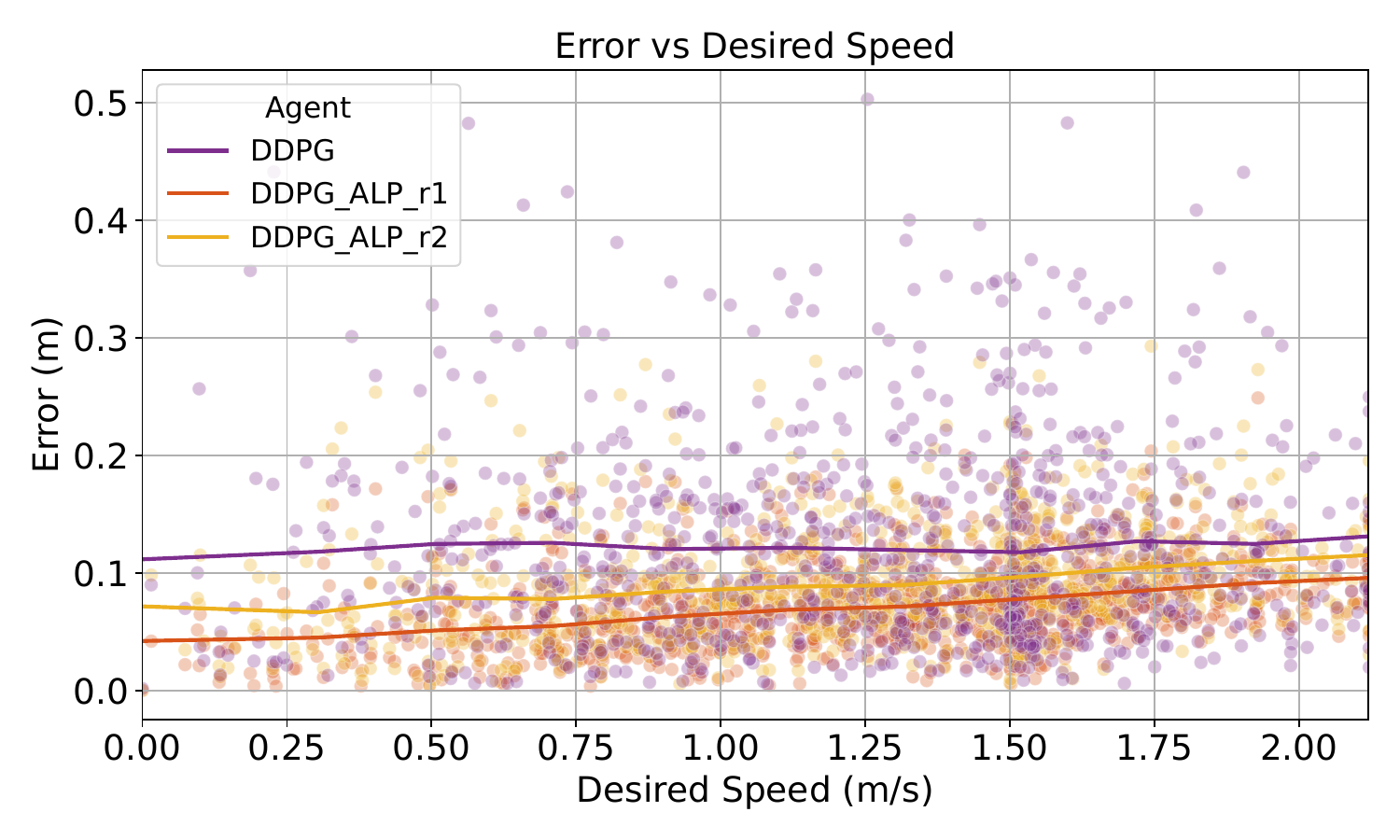}
        }
    }}
    \caption{Mean error between the target position and the end-effector position for the 4-cable robot over 10 episodes as a function of speed, using three different policies: DDPG with the conventional RL approach, DDPG with the ALP RL approach with reward~1, and DDPG with the ALP RL approach with reward~2.}
    \label{fig:comp_reward_config1}
\end{figure}

To analyze convergence time and maximum average return, we conducted training using the hyperparameters presented in the previous section, repeating the process four times with different random seeds. Each agent was trained using the same hyperparameters as the conventional reinforcement learning approach, with reward function~1 described in Section~\ref{sec:reward}.
Figure~\ref{fig:reward} shows that the PPO agent does not converge, while the DDPG and SAC agents both converge to good policies. The DDPG agent achieves a higher and more stable average reward than the SAC agent, as indicated by the lower variation in average reward. Therefore, we selected the DDPG agent for the remainder of the experiments due to its stability and reliable convergence.

\subsection{Testing results in simulation and comparison between reward functions}
We trained the ALP DDPG policy using both the collaborative reward function~1 and the individual reward function~2, as described in Section~\ref{sec:reward}, on the 4-cable robot (see Figure~\ref{fig:schemcablerobot}) for four different runs with random . The same procedure was applied to the DDPG with the conventional RL approach, using the same hyperparameters and reward function~1. For each run, we saved the best policy and used it for the testing phase. 
\begin{figure}[thpb]
    \centering
    \framebox{\parbox{3in}{
        \centering
        \resizebox{\linewidth}{!}{
            \fontsize{11pt}{11pt}\selectfont
            \includegraphics{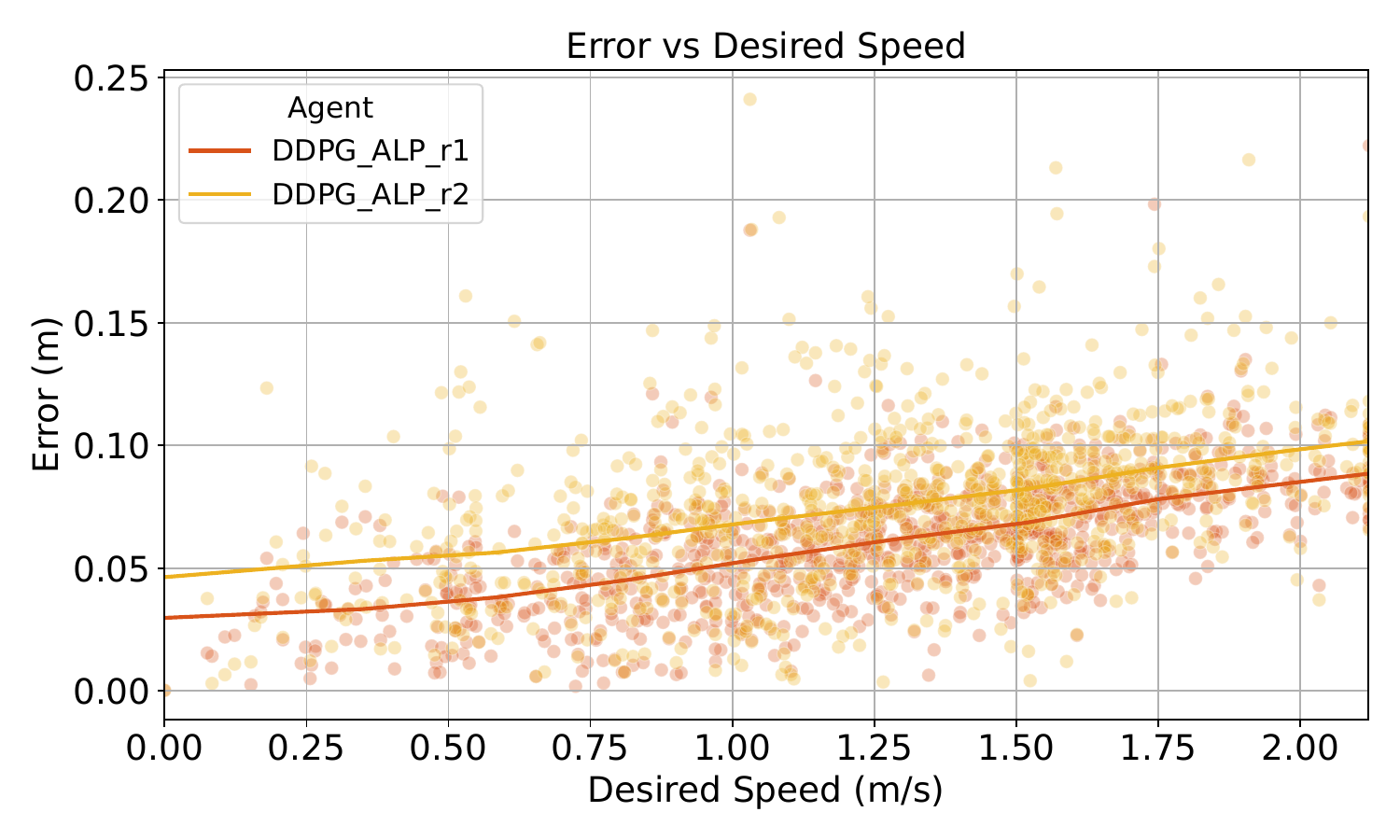}
        }
    }}
    \caption{Mean error between the target position and the end-effector position for the 4-cable robot configuration with $H_1=5$~m and $H_2=5$~m (see Figure~\ref{fig:schemcablerobot}) over 10 episodes as a function of speed, using the ALP RL approach with reward~1 (cooperative reward) and reward~2 (individual reward).}
    \label{fig:comp_reward_config2}
\end{figure}

We tested the three policies on the 4-cable robot with different speeds and target positions (10 episodes of 10 seconds each, as in training), using the same target positions for all policies. In Figure~\ref{fig:comp_reward_config1}, the error ranges from $0$ to $0.5$~m for the conventional RL approach and from $0$ to $0.25$~m for the ALP RL approach, indicating that the ALP RL outperforms the conventional RL approach. We also observe that the policy using reward~1 (cooperative reward) outperforms the policy using reward~2 (individual reward). This difference may be attributed to the reward coefficients not being optimally tuned for the individual reward, as the same coefficients were used for both reward functions, and were primarily tuned for the cooperative reward in the conventional RL approach.
\begin{figure}[thpb]
    \centering
    \framebox{\parbox{3in}{
        \centering
        \resizebox{\linewidth}{!}{
            \fontsize{11pt}{11pt}\selectfont
            \includegraphics{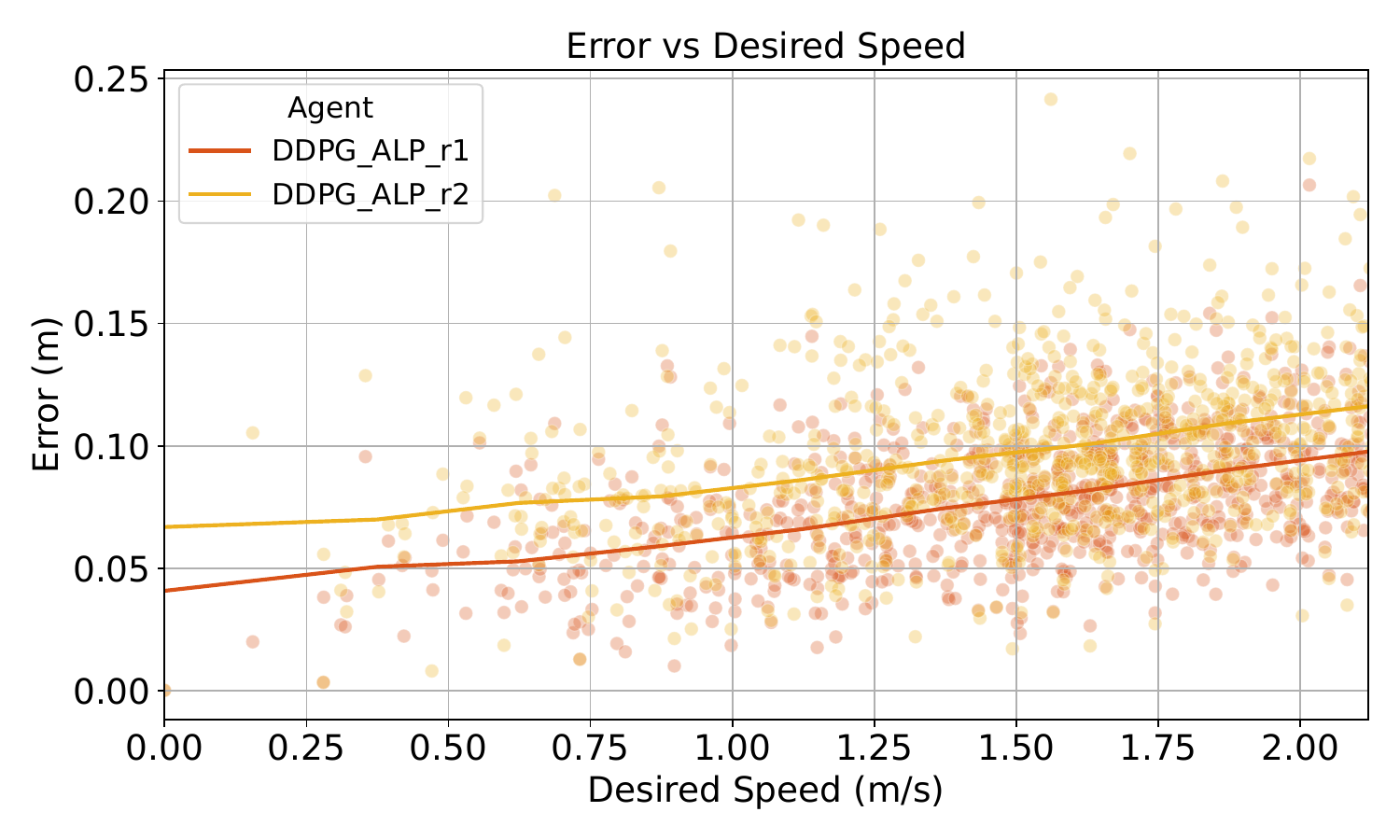}
        }
    }}
    \caption{Mean error between the target position and the end-effector position for the 8-cable robot configuration over 10 episodes as a function of speed, using the ALP RL approach with reward~1 (cooperative reward) and reward~2 (individual reward).}
    \label{fig:comp_reward_config3}
\end{figure}

However, the real advantage of the individual reward becomes apparent during fine-tuning, when additional training is performed in the implementation phase to achieve optimal performance. The cooperative reward can lead to unstable learning if the robot configuration differs from that used during training, as the scale of the cooperative reward may vary significantly between configurations. In contrast, the individual reward is more stable, since it depends only on the error of each actuator, making it suitable for fine-tuning across different configurations without issue.

Another important observation is that the performance of the ALP reinforcement learning approach surpasses that of the DDPG with the conventional RL approach. Thus, the ALP reinforcement learning approach outperforms the conventional RL approach, as it can be applied to different robot configurations with varying numbers of actuators without requiring additional training.

\subsection{Testing results on diffrent configurations}
We tested the three policies on the 4-cable robot with a different configuration (workspace with $H_1=5$~m and $H_2=5$~m, see Figure~\ref{fig:schemcablerobot}), using the same target positions for all policies. The conventional RL approach fails to track the desired trajectory when trained on a different configuration; it can only track trajectories in new configurations if the difference is small and the number of actuators remains the same.

In Figure~\ref{fig:comp_reward_config2}, we observe that the ALP reinforcement learning approach successfully tracks the desired trajectory in the new configuration, even though it was trained only on a 4-cable robot with a different setup. This is because the policy is learned to control each actuator independently, rather than the entire robot, allowing it to be applied to any configuration without additional training.
\begin{figure}[thpb]
    \centering
    \framebox{\parbox{3in}{
        \centering
        \resizebox{\linewidth}{!}{
            \fontsize{11pt}{11pt}\selectfont
            \includegraphics{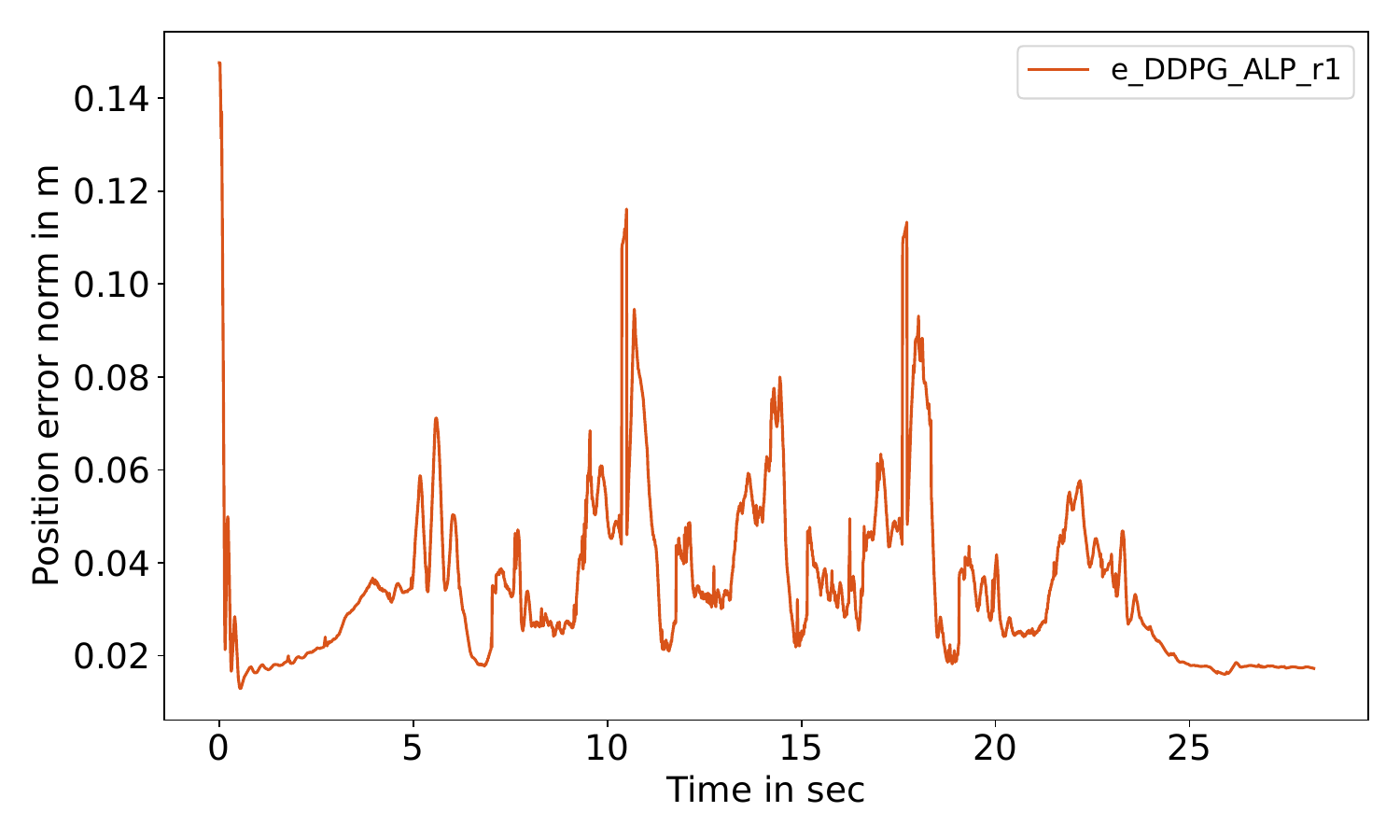}
        }
    }}
    \caption{Trajectory tracking of the real robot using the ALP reinforcement learning approach with reward~1. The error between the target position and the end-effector position is less than $0.04$~m.}
    \label{fig:real_robot_result}
\end{figure}

We also tested both policies on the 8-cable translational robot, where the robot has 8 actuators and the end-effector is a point mass, with all cables attached to the same point on the mobile platform. The same target positions were used for both policies. As shown in Figure~\ref{fig:comp_reward_config3}, the performance of both policies on this configuration is similar to their performance on the initial configuration, demonstrating the transferability of the policy learned on the 4-cable robot to the 8-cable robot.

\subsection{Testing results on the real robot}
We tested the best policy using reward~1 on the real robot in the 8-cable translational robot configuration. The policy's performance on the real robot closely matches its performance in simulation. As shown in Figure~\ref{fig:real_robot_result}, the agent is able to track the desired trajectory with high accuracy, and the error between the target position and the end-effector position is less than $0.04$~m.
\subsection{Robustness to errors in the position estimation}

\begin{table}[thpb]
    \centering
    \caption{A comparison of the performance of the different controllers}
    \label{table:performance}
    \normalsize
    \begin{tabular}{|c||c|c|c|}
        \hline
        Configuration $\backslash$ Controller & Classical & RL & ALP \\
        \hline
        4 cables                              & $\checkmark$ & $\checkmark$ & $\checkmark$ \\
        \makecell{Different mass than training\\($m=0.5,\dots,2$ kg)}          & $\checkmark$ & $\checkmark$ & $\checkmark$ \\
        \makecell{Other configurations \\ than training} & $\checkmark$ & $\checkmark$ & $\checkmark$ \\
        n cables                              & $\checkmark$ & $\times$     & $\checkmark$ \\
        Error in position estimation          & $\times$     & $\times$     & $\checkmark$ \\
        \hline
    \end{tabular}
\end{table}

Most classical controllers use the position of the end-effector to compute the control signal. In the absence of direct position measurement (i.e., no sensor for the end-effector position), the position is estimated using the robot's forward kinematics. Errors in position estimation can significantly degrade controller performance, especially if a cable measurement is incorrect or a cable is sagged, resulting in inaccurate end-effector position and poor trajectory tracking. 

In contrast, our ALP reinforcement learning approach does not rely on estimating the end-effector position. The control signal for each actuator is generated independently of the lengths of the other cables, allowing the system to remain robust even if there are errors in the measurement of one cable. As a result, the ALP RL controller can still track the desired trajectory under such conditions.

Table~\ref{table:performance} summarizes the major advantages of the ALP reinforcement learning approach compared to classical controllers.

\section{Conclusions}
In this work, we have proposed a new reinforcement learning-based control approach for cable-driven parallel robots using an actuator-level policy. This method enables systematic generalization, allowing control of any cable robot configuration. Simulation results demonstrate strong performance compared to conventional reinforcement learning approaches, and the proposed method has been validated on a real cable robot, confirming its transferability to different configurations and to real-world hardware.

\textbf{Limitations:} The comparison between different reinforcement learning algorithms remains inconclusive, mainly due to the manual tuning of hyperparameters, which significantly affects performance. Similarly, tuning the reward function coefficients is challenging and does not guarantee optimal results. Another limitation is that the CDPR configurations considered in this paper are restricted to translational CDPRs with a point mass mobile platform. The proposed approach could be extended to rotational CDPRs with a rigid body mobile platform; preliminary results are promising, but further work is needed to fully validate this extension.



\appendix

\section{Proof of concept}
\label{sec:proofconcept}
To demonstrate the feasibility of our approach, we use the dynamic model of a single cable.

\subsection*{Assumptions}
The following assumptions are made in deriving the dynamic model:
\begin{itemize}
    \item The cable is inextensible.
    \item The cable is wound around the winch without slipping.
    \item There is no dry or viscous friction acting on the system.
    \item The winch applies a torque \( \tau(t) \) which is converted to a tension \( T(t) \) at one end of the cable, proportional to the current \( I(t) \) flowing through the motor.
    \item The mobile platform applies a force \( F_{\text{ext}}(t) \) at the other end of the cable.
    \item The length of the cable between the winch and the end-effector is \( L(t) \).
    \item The mass of the cable between the winch and the end-effector is \( m(t) \).
\end{itemize}

\subsection*{Dynamic model of the cable}
Applying Newton's second law to the cable:
\begin{equation}
    m(t) \ddot{L}(t) = F_{\text{ext}}(t) - T(t)
    \label{eq:dynamicmodelcable}
\end{equation}

\subsection*{$F_{\text{ext}}(t)$ and $m(t)$ estimation}
Assuming the cable is inextensible and has uniform mass, \( m(t) \) can be estimated from the cable length and mass per unit length. \( F_{\text{ext}}(t) \) can be estimated from \eqref{eq:dynamicmodelcable} using the estimated \( m(t) \), the measured tension \( T(t) \) (proportional to the motor current \( I(t) \)), and the measured values of \( L(t) \) over recent time steps.

\subsection*{Existence of a control law}
Let \( \hat{F}_{\text{ext}} \) and \( \hat{m} \) be estimates of \( F_{\text{ext}} \) and \( m \), with:
\begin{align}
    \begin{split}
        \hat{F}_{\text{ext}}(t) &= F_{\text{ext}}(t) - \epsilon_{\text{ext}}(t) \\
        \hat{m}(t) &= m(t) - \epsilon_m(t)
    \end{split}
\end{align}
where \( \epsilon_{\text{ext}}(t) \) and \( \epsilon_m(t) \) are small bounded errors.

The objective is to track a desired cable length \( L_d(t) \), i.e., to choose \( T(t) \) such that:
\begin{align}
    \|L_d(t) - L(t)\| = \|e(t)\| \leq \epsilon_d
    \label{eq:connvergencecondition}
\end{align}
where \( \epsilon_d \) is small (or zero if estimation errors vanish).

Assume \( L(t) \) and its derivatives are bounded, which holds for the cable robot.

If we choose:
\begin{equation}
    T(t) = \hat{F}_{\text{ext}} - \hat{m} \left( \ddot{L}_d(t) + k_v\dot{e}(t) + k_p e(t) \right)
\end{equation}
then substituting into \eqref{eq:dynamicmodelcable} yields:
\begin{align}
    F_{\text{ext}} - \hat{F}_{\text{ext}} + \hat{m} \left( \ddot{L}_d(t) + k_v\dot{e}(t) + k_p e(t) \right) = m\ddot{L}(t) \\
    \epsilon_{\text{ext}}(t) + \hat{m} \left( \ddot{e}(t) + k_v\dot{e}(t) + k_p e(t) \right) = \epsilon_m \ddot{L}(t) \\
    \ddot{e}(t) + k_v\dot{e}(t) + k_p e(t) = \epsilon(t)
\end{align}
with
\[
    \epsilon(t) = \frac{\epsilon_m \ddot{L}(t) - \epsilon_{\text{ext}}(t)}{\hat{m}}
\]

This is a second-order equation, and by choosing \( k_v > 0 \) and \( k_p > 0 \), we ensure convergence as in \eqref{eq:connvergencecondition}. The choice of \( k_v \) and \( k_p \) can also be formulated as a convex optimization problem.

Thus, this control law can be learned by a neural network using a reinforcement learning agent, provided the actuator state includes the cable length at previous time steps (to estimate cable speed and tension at the other end) and the motor current (to estimate the force applied to the cable).

\section{Current constraints}
\label{sec:currentconstraints}

\subsection*{Current control loop}
Cable-driven parallel robots use cables to transmit forces from the motors to the end-effector. The control signal is typically the cable tension, but since tension is not directly measurable, the motor current is used instead. The CDPR in our lab employs a proportional control loop where the motor current is proportional to the difference between the desired speed and the measured speed of the motors:
\[
    i = K_{motor}(u - v_{m})
\]
where $i$ is the motor current, $K_{motor}$ is the motor constant, $u$ is the desired motor speed, and $v_{m}$ is the measured motor speed.

From this control loop, any desired speed less than the measured speed results in a negative desired current, meaning the motor rotates in the opposite direction and cable tension drops to zero, which is unsafe. While we typically control the desired speed, having access to the measured speed allows us to adjust the desired speed to respect current limits.

\subsection*{Current bounds}
Cables in a CDPR must always remain in tension; slack cables risk entanglement and loss of robot mobility, requiring intervention to restore safety. To prevent this, we set a minimum current limit $i_{min}$. Conversely, the current must not exceed a maximum limit $i_{max}$ to avoid damaging the motors, cables, or end-effector. Thus, we enforce:
\[
    i_{sat} = 
    \begin{cases}
        i_{max} & \text{if } i > i_{max} \\
        i       & \text{if } i_{min} < i < i_{max} \\
        i_{min} & \text{if } i < i_{min}
    \end{cases}
    \label{isat}
\]
Based on the current control loop, the agent's action is computed as:
\[
    u_{sat} = \frac{i_{sat}}{K_{motor}} + v_{m}
    \label{usat}
\]

By computing the agent's action this way, we ensure the motor current remains within safe limits and cables stay in tension. Mapping unsafe actions to the nearest safe action ensures the agent does not select unsafe commands and learns to control the robot across all states of the environment.
\section{Desired trajectory generation: the case of trajectory tracking in bounded workspace}
\label{sec:targetgen}
\begin{table}[thpb]
    \centering
    \caption{Target generation parameters}
    \label{tb:targetgen}
    \normalsize
    \begin{tabular}{|c|c|}
        \hline
        Parameter                      & Value               \\ \hline
        Maximum acceleration in x-axis & 35 $\mathrm{m/s^2}$ \\ 
        Maximum acceleration in z-axis & 35 $\mathrm{m/s^2}$ \\ 
        Maximum velocity in x-axis     & 4 $\mathrm{m/s}$    \\ 
        Maximum velocity in z-axis     & 4 $\mathrm{m/s}$    \\ \hline
    \end{tabular}
\end{table}

The trajectories are generated as follows: 
First, we set the maximum acceleration $a_{max}$ and the maximum speed $v_{max}$. 
A random initial position $X_0$ is chosen at the start of each episode, with speed and acceleration reset to zero. At each time step $t$, a random acceleration $a_t$ is generated from a Gaussian distribution such that $-a_{max} \leq a_t \leq a_{max}$. The speed $v_t$ is then updated by integrating the acceleration:
\[
    v_t = v_{t-1} + a_t \cdot dt
\]
If the speed exceeds the limits, it is clipped to the boundary value. The trajectory is then generated by updating the position:
\[
    X_t = X_{t-1} + v_t \cdot dt
\]


This approach works very well for low speeds or in an unbounded workspace, but at high speeds, the end-effector often reaches the workspace limits before the episode ends. To address this, we initially tried reducing the episode length, but this led to instability due to fewer reward samples for estimating the average return.

Instead of shortening the episode, we chose to randomly reverse the desired speed $v_{t}$ when the end-effector reaches the workspace boundaries, keeping it within the workspace. While this method works for high speeds, it is not ideal, as the resulting trajectories may not be smooth or fully consistent with the robot's dynamic limits. However, it is acceptable during training because it ensures the agent explores all states of the environment and learns to control the robot under various conditions.

We use the parameters shown in Table~\ref{tb:targetgen} to generate trajectories for our CDPR. The velocity limits are set to match the robot's capabilities, with the nominal motor speed at $3000~\mathrm{rpm}$, equivalent to $4~\mathrm{m/s}$ for the cables. The acceleration limits are set higher than the robot's maximum acceleration to allow reaching maximum speed within the limited episode length, and because acceleration is sampled from a random Gaussian distribution, the robot does not reach maximum acceleration in all states.

\section{Hyperparameters and CDPR parameters}
\label{sec:hyperparams}
The training process is performed in simulation using the OpenAI Gym environment. The CDPR used is a 4-cable robot with parameters listed in Table~\ref{tb:CDPRparams}.
\begin{table}[thpb]
    \centering
    \normalsize
    \vspace{1em}
    \caption{CDPR parameters}
    \label{tb:CDPRparams}
    \begin{tabular}{|c|c|c|}
        \hline
        Parameter                       & Symbol        & Value                 \\ \hline
        Nominal current                 & $I_{nom}$     & 6.8 $\mathrm{A}$      \\
        Nominal speed                   & $V_{nom}$     & 3420 $\mathrm{rpm}$   \\
        Nominal torque                  & $T_{nom}$     & 0.8 $\mathrm{Nm}$     \\
        Control signal frequency        & $f_e$         & 100 $\mathrm{Hz}$     \\
        Motor torque constant           & $K_c$         & 0.123 $\mathrm{Nm/A}$ \\
        \makecell{Motor current \\ proportional gain} & $K_{motor}$   & 1 $\mathrm{A/rps}$    \\
        Drum radius                     & $r$           & 0.0178 $\mathrm{m}$   \\
        End-effector mass               & $m$           & 1 $\mathrm{kg}$       \\ 
        Time step                       & $dt$          & 0.01 $\mathrm{s}$     \\
        \hline
    \end{tabular}
\end{table}

For each agent, a manual hyperparameter tuning process was performed to achieve the best average return and fastest convergence. The hyperparameters used for each agent are presented in Table~\ref{tb:Hyperparameters}. The hyperparameters were tuned to achieve the best performance for the 4-cable robot configuration, and the same hyperparameters were used for all other configurations.
 \begin{table}[htpb]
            \centering
            \normalsize
            \vspace{1em}
            \caption{Hyperparameters}\label{tb:Hyperparameters}
            \begin{tabular}{cccc}
                Hyperparameter & DDPG&PPO&SAC \\
                \hline
                \makecell{Max number\\ of episodes} & $3 \cdot10^{4}$ & $8 \cdot10^{4}$& $3 \cdot10^{4}$ \\
                \makecell{Max number\\ of steps} & 1000 & 1000 & 1000 \\
                Q learning rate & $4\cdot10^{-5}$ & -& $3\cdot10^{-4}$ \\
                Value learning rate & - & $1\cdot10^{-5}$&$3\cdot10^{-4}$\\
                Actor learning rate & $2\cdot10^{-5}$ & $2\cdot10^{-5}$ & $3\cdot10^{-4}$ \\
                Update rate & $5\cdot10^{-4}$& $5\cdot10^{-4}$ & $5\cdot10^{-4}$ \\
                Discount factor & 0.99 & 0.99 &0.99\\
                Buffer size & 50000& - & 50000 \\
                Batch size & 128 & 50 & 128 \\
                Noise deviation& 0.033 & - & - \\
                Alpha coeff.& - & - & 0.01 \\
                GAE & - & 0.95 & - \\
                Log std dev & - & -1.3 & - \\
                Epoch episodes & - & 10 & - \\
                Iterations & - & 4 & - \\
                Hidden layers & 2 & 2 & 2 \\
                Hidden units & 128 & 128 & 128 \\
                \hline
            \end{tabular}
    \end{table}
The training time depends on available computational resources and the experiment setup. During hyperparameter tuning, training is stopped when the agent's average return stabilizes; for the final training, a fixed number of episodes is used to allow the agent to learn the optimal policy for the task. Table~\ref{tb:Training_time} presents the training time per episode for each agent.

 \begin{table}[htpb]
    \begin{center}
            \normalsize
            \caption{Training Time}\label{tb:Training_time}
            \begin{tabular}{cccc}
                    Agent&   \makecell{Training \\time per \\episode}                   & \makecell{Number \\of episodes}& \makecell{Training\\ time per \\ experiment}                        \\
                    \hline
                    \makecell{DDPG \\ SAC} & $6-10(s)$ & $1-3\cdot10^{4}$ & $16-85(h)$ \\
                    PPO & $4.5-8(s)$ & $4-8\cdot10^{4}$ & $50-100(h)$ \\
                    \hline
            \end{tabular}
    \end{center}
 \end{table}



\bibliographystyle{elsarticle-num-names}\bibliography{mybibfile.bib}

\end{document}